%% file: main.tex
\RequirePackage{amsmath}
\documentclass[11pt]{article}
\usepackage[margin=2cm,a4paper]{geometry}
\usepackage[bitstream-charter]{mathdesign}
\usepackage{soul}
\usepackage{url}
\usepackage[hidelinks]{hyperref}
\usepackage[utf8]{inputenc}
\usepackage{graphicx}
\usepackage{booktabs}
\usepackage{latexsym}
\usepackage{rotating}
\usepackage{breakcites}

\let\snake\leadsto

\allowdisplaybreaks

\newcommand{\proba}{\textbf{\textsf{\textup{p}}}} 
\usepackage{xcolor}

\begin{document}

\title{From Shallow to Deep Interactions Between Knowledge Representation, Reasoning and Machine Learning}
\author{Kay R. Amel$^1$\\GDR Aspects Formels et Algorithmiques de l'Intelligence Artificielle\\CNRS}
\maketitle
\begin{abstract}
This paper\footnotetext[1]{Kay R. Amel is the pen name of the working group ``Apprentissage et Raisonnement'' of the GDR (``Groupement De Recherche'') named ``Aspects Formels et Algorithmiques de l'Intelligence Artificielle'', CNRS, France (https://www.gdria.fr/presentation/). The contributors to this paper include: Zied Bouraoui (CRIL, Lens, Fr, bouraoui@cril.fr), Antoine Cornuéjols (AgroParisTech, Paris,Fr, antoine.cornuejols@agroparistech.fr), Thierry Den{\oe}ux (Heudiasyc, Compiègne, Fr, thierry.denoeux@utc.fr), Sébastien Destercke (Heudiasyc, Compiègne, Fr, sebastien.destercke@hds.utc.fr), Didier Dubois (IRIT, Toulouse, Fr, dubois@irit.fr), Romain Guillaume (IRIT, Toulouse, Fr, Romain.Guillaume@irit.fr), Jo\~ao Marques-Silva (ANITI, Toulouse, Fr, joao.marques-silva@univ-toulouse.fr), Jérôme Mengin (IRIT, Toulouse, Fr, Jerome.Mengin@irit.fr), Henri Prade (IRIT, Toulouse, Fr, prade@irit.fr), Steven Schockaert (School of Computer Science and Informatics, Cardiff, UK, SchockaertS1@cardiff.ac.uk), Mathieu Serrurier (IRIT, Toulouse, Fr, mathieu.serrurier@gmail.com), Christel Vrain (LIFO, Orléans,Fr, Christel.Vrain@univ-orleans.fr).} proposes a tentative and original survey of meeting points between Knowledge Representation and Reasoning (KRR) and Machine Learning (ML), two areas which have been developing quite separately in the last three decades. Some common concerns are identified and discussed such as the types of {used representation}, the roles of knowledge and data, the lack or the excess of information, {or} the need for explanations and causal understanding. Then some methodologies combining reasoning and learning are reviewed (such as inductive logic programming, neuro-symbolic reasoning, formal concept analysis, rule-based representations and ML, uncertainty in ML, or case-based reasoning and analogical reasoning), before discussing examples of synergies between KRR and ML (including topics such as belief functions on regression, EM algorithm versus revision, the semantic description of vector representations, the combination of deep learning with high level inference, knowledge graph completion, declarative frameworks for data mining, or preferences and recommendation). This paper is the first step of a work in progress aiming at a better mutual understanding of research in KRR and ML, and how they could cooperate.
\end{abstract}

\section{Introduction}

Reasoning and learning are two basic concerns at the core of Artificial Intelligence (AI). In the last three decades, Knowledge Representation and Reasoning (KRR) on the one hand, and Machine Learning (ML) on the other hand, have been considerably developed and have specialised themselves in a large number of dedicated sub-fields. These technical developments and specialisations, while they were strengthening the respective corpora of methods in KRR and in ML, also contributed to an almost complete separation of the lines of research in these two areas, 
making many researchers on one side largely ignorant of 
{what is going on, on the other side.}

This state of affairs is also somewhat relying on general, overly simplistic, dichotomies that suggest there exists a large gap between KRR and ML: KRR deals with knowledge, ML handles data; KRR privileges symbolic, discrete approaches, while numerical methods dominate ML. Even if such a rough picture points out facts that cannot be fully denied, it is also misleading, as for instance KRR can deal with data 
as well \cite{Pra2016} (e.g., in formal concept analysis) and ML approaches may rely 
on symbolic knowledge (e.g., {in} inductive logic programming). Indeed, the frontier between the two fields is actually much blurrier than it appears, as both are involved in approaches such as Bayesian networks, or 
case-based reasoning and analogical reasoning, and they share important concerns such as 
uncertainty representation (e.g., probabilistic or possibilistic models, belief functions, imprecise probability-based approaches). 

These remarks already suggest that KRR and ML may have more in common than one might think at first glance. In that respect, it is also important to remember that the human mind is able to perform both reasoning and learning tasks with many interactions between these two types of activity. In fact, from the very beginning of the AI history, both reasoning and learning tasks have been considered, but not by the same researchers;
see, e.g., \cite{DP19}. 
So, especially if the ultimate goal of AI is to have machines {that} perform tasks handled by the human mind, it might be natural and useful to increase the cooperation between KRR and ML. 

The intended goal pursued in this work is to start and construct an inventory of common concerns in KRR and ML, of methodologies combining reasoning principles and learning, of examples of KRR/ML synergies. Yet, this paper is not an overview of the main issues in KRR crossed with an overview of the main issues in ML, trying to identify when they meet. Doing so would lead to a huge and endless survey since providing a survey of methodologies and tools for KRR alone, or for ML alone would be already a colossal task$^2$\footnotetext[2]{Nevertheless the interested reader is referred to appropriate chapters in \cite{GuidedTour} or to monographs such as \cite{BraLev,Baral,halpern2017reasoning,UML,FML}.}. In the following we rather try to identify a collection of meeting points between KRR and ML. Since it is a work in progress, we do not expect to reach any form of exhaustiveness, and even {some} important topics may remain absent from the document at this stage.

The paper is not either an introductory paper to KRR {and/or} ML. It is rather intended for readers 
{who are quite familiar with}
 either KRR or ML, and who are curious about the other field. It aims in the long range at contributing to a better mutual understanding of the two communities, and maybe to identify some synergies worth of further research combining KRR and ML. 



\section{Common Concerns}
In order to suggest and illustrate differences
{and} also similarities between KRR and ML, let us start with the simple example of a classification or recommendation-like task 
,such as, e.g., 
associating the profile of a candidate (in terms of skills, tastes) with possible activities suitable for him/her in a vocational guidance system. Such a problem may be envisioned in different manners. On the one hand, one may think of it in terms of a rule-based system relying on some expertise (where rules may be pervaded with uncertainty), or on the other hand in terms of machine learning by exploiting a collection of data (here pertaining to past cases in career guidance). 

It is worth noticing that beyond the differences of types of representation that are used in both kinds of approach (e.g., conditional tables for uncertainty assessment vs. weights in a neural net), there are some noticeable similarities between (graphical) structures that can be associated with a rule-based reasoning device, handling uncertainty (or an information fusion process) and with a neural net. This remark suggests that, beyond differences in perspective, there is some structural resemblance between the two types of process. This resemblance has been investigated recently in detail in the setting of belief function theory \cite{Denoeux19}, but an example may also be found in an older work on a possibilistic ($\max$-$\min$) matrix calculus devoted to explainability (where each matrix represents a rule) \cite{FarPra89}.

Beyond this kind of parallel, it is clear that KRR and ML have common concerns. This section 
{gives an overview of} the main ones regarding {the} representation issues, {the} complexity, the role of knowledge, the handling of lack of information, or information in excess, uncertainty, and last but not least regarding causality and explanation. Each subsection below tries to follow the same basic structure, by each time providing
i) the KRR view, ii) the ML view, and iii) some 
synthesis and discussion. 
\subsection{{Types of Representation}}
In KRR, as suggested by the name, the main representations issues concern the representation of pieces of knowledge (rather than data).
{The large variety of real world information} has led to 
{a number of} logical 
{formalisms}
ranging from classical logic (especially proportional and first order) to modal logics (for dealing with e.g., time, deontic, or epistemic notions) and to non classical logics for handling commonsense reasoning.

The representation may use different formats, directed or undirected: sets of if-then rules, or sets of logical formulas. 
A rule ``if $A$ then $B$'' is a 3-valued entity (as first noticed in \cite{DeF36}), since it induces a partition between its set of examples, its set of counterexamples and the set of items for which the rule is irrelevant (i.e., when $A$ is false). So a rule strongly departs from its apparent logical counterpart in terms of material implication $A \to B$ (which is indeed non-directed, since equivalent $\neg B \to \neg A$). This discrepancy can be also observed in the probabilistic setting, since 
$Prob(B|A) \neq Prob(A \to B)$ in general. Rules may hold up to (implicit) exceptions (see subsection \ref{lackexcess}).

Knowledge may be pervaded with uncertainty, which can be handled in different settings, in terms of probability, possibility, belief functions, or imprecise probabilities (see subsection \ref{lackexcess}). In all of these cases, a joint distribution can be decomposed in sub-distributions laying bare some form of conditional independence relations, with a graphical counterpart; the prototypical graphical models in each representation are respectively Bayesian networks (probabilistic), possibilistic networks, credal networks (imprecise probabilities \cite{Cozman00}) or valuation-based systems (belief functions). Conceptual graphs \cite{sowa,cheinmugnier} offer a graph representation for logic, especially for {ontologies/description logics. }

The main goal of KRR is to develop sound and (as far as possible complete) inference mechanisms to draw conclusions from generic knowledge and factual data, in a given representation setting \cite{halpernetal,halpern2017reasoning}. The mathematical tools underlying KRR are those of logic and uncertainty theories, and more generally discrete mathematics. 
An important issue in KRR is to find good compromises between the expressivity of the representation and {the} computational tractability for inferring the conclusions of interest {from it} \cite{LevesqueB87}. This concern is especially at work with description logics that are bound to use tractable fragments of first order logic. 

The situation in ML is quite different concerning representation issues. ML aims at learning a model of the world from data. There are thus two key representation problems{:} the representation of data and the representation of models. See, e.g., \cite{CorKorNoc,CorVrain}. 
In many approaches the data space is assimilated to a subset of ${\mathbb R}^p$, in which the observations are described by $p$ numerical attributes. This is the simplest case, allowing the use of mathematical results in linear algebra and in continuous optimization. Nevertheless data may also be described by qualitative attributes, as for instance binary attributes, thus requiring different mathematical approaches, based on discrete optimisation and on enumeration coupled with efficient pruning strategies. Quite often, data is described by both types of attributes and only few ML tools, for instance decision trees, are able to handle them. Therefore, changes of representation are needed, as for instance discretization, or the encoding of qualitative attributes into numerical ones, all inducing a bias on the learning process. More complex data, such as relational data, trees, graphs need more powerful representation languages, such as first order logic or some proper representation trick as for instance propositionalization or the definition of appropriate kernels. It is important to notice that the more sophisticated the representation language, the more complex the inference process and a trade-off must be found between the granularity of the representation and the efficiency of the ML tool.

Regarding models, they depend on the ML task: supervised or unsupervised classification, learning to rank, mining frequent patterns, etc. They depend also on the type of approach that one favours: more statistically or more artificial-intelligence oriented. There is thus a distinction between \textit{generative} and \textit{discriminative} models (or decision functions). In the \textit{generative approach}, one tries to learn a probability distribution $\proba_{\cal X}$ over the input space $\cal X$. If learning a precise enough probability distribution is successful, it becomes possible in principle to generate further examples ${\mathbf x} \in {\cal X}$, the distribution of which is indistinguishable from the true underlying distribution. It is sometimes claimed that this capability makes the generative approach ``explicative''. This is a matter of debate. The \textit{discriminative} approach does not try to learn a model that allows the generation of more examples. It only provides either a means of deciding when in the supervised mode, or a means to express some regularities in the data set in the unsupervised mode. These regularities as well as these decision functions can be expressed in terms of logical rules, graphs, neural networks, etc. While they do not allow to generate new examples, they nonetheless can be much more interpretable than probability distributions.

Very sketchily, one can distinguish between the following types of representations.

\begin{itemize}
 \item Linear models and their generalisations, such as linear regression or the linear perceptron first proposed by Rosenblatt \cite{R58}. Because these models are based on linear weightings of the descriptors of the entries, it looks easy to estimate the importance of each descriptor and thus to offer some understanding of the phenomenon at hand. This, however, assumes that the descriptors are uncorrelated and are well chosen.
 \item Nonlinear models are often necessary in order to account for the intricacies of the world. Neural networks, nowadays involving very numerous layers of non linearity, are presently the favourite tools for representing and learning non linear models. 
 \item Linear models as well as nonlinear ones provide a description of the world or of decision rules through (finite) combinations of descriptors. They are parametric models. Another approach is to approximate the world by learning a non previously fixed number of prototypes and use a nearest-neighbour technique to define decision functions. These systems are capable of handling any number of prototypes as long as the can fit the data appropriately. \textit{Support Vector Machines} (SVM) fall in this category since they adjust the number of support vectors (learning examples) in order to fit the data. Here, explaining a rule may mean providing a list of the most relevant prototypes that the rule uses. 
 \item The above models are generally numerical in essence, and the associated learning mechanisms most often rely on some optimisation process over the space of parameters. Another class of models relies on logical descriptions, e.g., sets of clauses. Decision trees can also be considered as logic-based, since each tree can be transformed into a set of clauses. The learning algorithms use more powerful structures over the space of models than numerical models. In many cases the discrete nature of the search space and the definition of a generality relation between formulas allow the organization of models in a lattice and the design of heuristics to efficiently prune the search space. More generally, these approaches are usually modeled as enumeration problems (e.g., pattern mining) or discrete optimization problems (supervised learning, clustering, \ldots). Moreover such models offer more opportunities to influence the learning process using prior knowledge. Finally, they can be easily interpreted. The downside is their increased brittleness when coping with noisy data. 
\end{itemize}

\subsection{{Computational Complexity}}
Complexity issues are a major concern in any branch of computer science. In KRR, very expressive representation languages have been studied, but interesting reasoning problems for these languages are often at least at the second level of the polynomial hierarchy for time complexity. There is a trade-off between the expressive power of a language and the complexity of the inference it allows. Reasoning tasks in languages with suitably restricted expressivity are tractable, like for instance languages using Horn clauses or Lightweight description logics such as DL-lite~\cite{Calvaneseetal:aaai05} or EL~\cite{baader2005pushing}.

The study of complexity has motivated a large number of works in many fields of KRR including non-monotonic reasoning, argumentation, belief merging and uncertainty management. In particular when the desirable solution (i.e., gold standard) of the problem (for instance, merging operator, inconsistency-tolerant consequence relation, etc.) has a high computational complexity, then it is common to look for an approximation that has reasonable complexity. For instance, the observation that answering meaningful queries from an inconsistent DL-Lite knowledge base using universal consequence relation is NP-Complete, has led to the introduction of several tractable approximations \cite{DBLP:conf/jelia/BagetBBCMPRT16}.

The attempt to cope with hardness of inference has also been a driving force in research around some important and expressive languages, including propositional clauses and CSPs, where inference is NP-complete; for instance, powerful methods nowadays enable the solving of SAT problems with up to hundreds of thousands of variables, and millions of clauses in a few minutes (see section~\ref{sect:sat}). Some of the most competitive current SAT solvers are described in \cite{AbrameHabet:j-sat-bool-comp15,MartinsManquinhoLynce:sat14,LuoCaiWuJieSu:ieee-trans-comp15}. Two other ways to cope with time complexity are anytime methods, which can be interrupted at any time during the solving process and then return an incomplete, possibly false or sub-optimal solution; and approximate methods. A recent trend in KRR is to study so-called \textit{compilation schemes}~\cite{DarwicheMarquis:jair02,Marquis:aaai15}: the idea here is to pre-process some pieces of the available information in order to improve the computational efficiency (especially, the time complexity) of some tasks; this pre-processing leads to a representation in a language where reasoning tasks can be performed in polynomial time (at the cost of a theoretical blow up in worst-case space complexity, which fortunately does not often happen in practice).

Contrastingly, ML algorithms often have a time complexity which is polynomial in the number of variables, the size of the dataset and the size of the model being learnt, especially when the domains are continuous. However, because of the possible huge size of the dataset or of the models, capping the degree of the polynomial remains an important issue. In the case of discrete domains, finding the optimal model,i.e., the one that best fits a given set of examples, can be hard (see \cite{HyafilRivest:info-proc-let76} ), but one is often happy with finding a ``good enough'' model in polynomial time: there is no absolute guarantee that the model that best fits the examples is the target model anyway, since this may depend on the set of examples. In fact, an important aspect of complexity in ML concerns the prediction of the quality of the model that one can learn from a given dataset: in the PAC setting for instance \cite{Valiant:comm-acm84}, one tries to estimate how many examples are needed to guarantee that the model learnt will be, with a high probability, a close approximation to the unknown target model. Intuitively, the more expressive the hypothesis space is, the more difficult it will be to correctly identify the target model, and the more examples will be needed for that \cite{VapnikChervonenkis:th-proba-appli71}.

\subsection{Lack and Excess of Information: Uncertainty}\label{lackexcess}
		
With respect to a considered reasoning or decision task, information may be missing, or, on the contrary, may be in excess, hence in conflict, which possibly generates uncertainty.
Uncertainty has always been an important topic in KRR~\cite{halpern2017reasoning}. While in ML uncertainty is almost always considered to be of statistical or probabilistic origin (aleatory uncertainty), other causes for uncertainty exist, such as the sheer lack of knowledge, and the excess of information leading to conflicts (epistemic uncertainty). 
However, the role of uncertainty handling in KRR and in ML seems to have been very different so far. While it has been an important issue in KRR and has generated a lot of novel contributions beyond classical logic and probability, it has been considered almost only from a purely statistical point of view in ML~\cite{vapnik2013nature}. 

The handling of uncertainty in KRR has a long history, as much with the handling of incomplete information in non-monotonic reasoning as with the handling of probabilities in Bayesian nets \cite{Pearl:88}, and in probabilistic logic languages \cite{russell,cozman20}. Other settings that focus on uncertainty due to incomplete information are possibility theory, with weighted logic bases (possibilistic logic
\cite{DLP94,DuboisPS17}) and graphical representations (possibilistic nets
\cite{BenferhatDGP02,BenferhatT12}). Belief functions also lend themselves to graphical representations (valuation networks \cite{Shenoy94}, evidential networks \cite{YaghlaneM08}) and imprecise probability as well (credal nets \cite{Cozman05}).

Uncertainty theories distinct from standard probability theory, such as possibility theory or evidence theory are now well-recognised in knowledge representation. They offer complementary views to uncertainty with respect to probability, or as generalisations of it, dedicated to epistemic uncertainty when information is imprecise or partly missing. 


In KRR, at a more symbolic level, the inevitability of partial information has motivated the need for exception-tolerant reasoning. 
For instance, one may provisionally conclude that “Tweety flies” while only knowing that "Tweety is a bird”, 
although the default rule “birds fly" has exceptions, and we may later conclude that “Tweety does not fly”, when getting more (factual) information about Tweety.
Thus non-monotonic reasoning \cite{BMT2011} has been developed for handling situations with incomplete data, where only plausible tentative conclusions can be derived.
Generic knowledge may be missing as well. For example, one may not have the appropriate pieces of knowledge for concluding about some set of facts. Then it may call for interpolation between rules \cite{SchockaertP13}.

When information is in excess in KRR, it may mean that it is just redundant, but it becomes more likely that some inconsistency appears. Redundancy is not always a burden, and may sometimes be an advantage by making more things explicit in different formats (e.g., when looking for solutions to a set of constraints). 

Inconsistency is a natural phenomenon in particular when trying to use information coming from different sources. 
 Reasoning from inconsistent information is not possible in classical logic (without trivialisation). It has been extensively studied in AI \cite{BeHu98,BenferhatDP97,CC16}, in order to try and salvage non-trivial conclusions not involved in contradictions. Inconsistency usually appears at the factual level, for instance a logical base with no model. However, a set of rules may be said to be \emph{incoherent} when there exists an input fact that, together with the rules, would create inconsistency \cite{AyelRousset}.

Machine Learning can face several types of situations regarding the amount of information available. It must be said at once that induction, that goes from observations to regularities, is subject to the same kind of conservation law as in Physics. The information extracted is not created, it is just a reformulation, often with loss, of the incoming information. 

If the input data is scarce, then prior knowledge, in one form or another, must complete it. The less data is available, the more prior knowledge is needed to focus the search of regularities by the learning system. This is in essence what the statistical theory of learning says \cite{vapnik2013nature}. In recent years, lots of methods have been developed to confront the case where data is scarce and the search space for regularities is gigantic, specially when the number of descriptors is large, often in the thousands or more. The idea is to express special constraints in the so-called regularization term in the inductive criterion that the system use to search the hypothesis space. For instance, a constraint is often that the hypothesis should use a very limited set of descriptors \cite{tibshirani1996regression}. 

When there is plenty of data, the problem is more one of dealing with potential inconsistencies. However, except in the symbolic machine learning methods, mostly studied in the 1980s, there is no systematic or principled ways of dealing with inconsistent data. Either the data is pre-processed in order to remove these inconsistencies, and this means having the appropriate prior knowledge to do so, or one relies on the hope that the learning method is robust enough to these inconsistencies and can somehow smooth them up. Too much data may also call for trying to identify a subset of representative data (a relevant sample), as sometimes done in case-based reasoning, when removing redundant cases. Regarding the lack of data there is a variety of approaches for the imputation of missing values ranging from the EM algorithm \cite{Dempster77} to analogical proportion-based inference \cite{BouPraRicIJAR2017b}. However these methods get rid of incompleteness and do not reason about uncertainty.

Finally, a situation that is increasingly encountered is that of multi-source data. Then, the characteristics of the multiple data sets can vary, both in the format, the certainty, the precision, and so on. Techniques like data fusion, data aggregation or data integration are called for, often resorting again to prior knowledge, using for instance ontologies to enrich the data.

\subsection{Causality and Explainability} \label{causex}
"What is an explanation", "What has to be explained, and how" are issues that have been discussed for a long time by psychologists and philosophers \cite{Thagard78,Bromberger92}.
The interest in AI for explanations is not new either. It appears with the development of rule-based expert systems in the mid-1980's. Then there was a natural need for explanations that are synthetic, informative, and understandable for the user of an expert system \cite{CDL19}. 
This raises issues such as designing strategic explanations for a diagnosis, for example in order to try to lay bare the plans and methods used in reaching a goal \cite{HaslingCR84}, or 
using ``deep'' knowledge for improving explanations
\cite{Kassel87}. Another issue was the ability to provide negative explanations (as well as positive ones) for answering questions of the form "Why did you not conclude X?" \cite{RoussetS87}, even in the presence of uncertainty \cite{FarPra89}.

Introductory surveys about explanations in AI may 
be found in a series of recent papers \cite{HoffmanK17a,klein-ieee-is17b,klein-ieee-is18a,klein-ieee-is18b,GilpinBYBSK18,corrGilpin}. Let us also mention the problem of explaining the results of a multi-attribute preference model that is like a ``black box''. It has been more recently studied in \cite{Labreuche11}.

As now discussed, explanations are often related to 
the idea of causality \cite{HalpernPearl05}. Indeed most of the explanations we produce or we expect involve some causal relationships (e.g., John imposed on himself to go to the party \textit{because} he thought that Mary would be there). In many domains where machines can provide aid for decision making, as in medicine, court decisions, credit approval and so on, decision makers and regulators more and more want to know what is the basis for the decision suggested by the machine, why it should be made, and what alternative decision could have been made, had the situation been slightly different. 
One of the difficulties of explainability in Machine learning is due to the fact that algorithms focus on correlations between features and output variables rather than on causality. The example of {\em wolf vs dog} identification \cite{guestrin-kdd16} perfectly illustrates this problem. When using a deep classifier, the main feature that determines if a picture represents a dog or a wolf is the presence of snow. There obviously exists a correlation between snow and wolves but it is clearly not a causality link. When this unwanted bias is known, this can be corrected by adding constraint or balancing the dataset. However, due to the lack of interpretability of some algorithms, identifying these biases is challenging. On the other hand, the problem of constraining an algorithm to learn causality rather than correlation is still open. 

If explainability is quickly becoming a hot topic in ML, while it was such for expert systems about 30 years ago, the search for solutions is still at an early stage. Some tentative distinction between \textit{interpretability} and \textit{explainability} has been suggested. 

Interpretability may have a statistical or a KR interpretation (which are of course not mutually exclusive). From a statistical point of view, an interpretable model is a model that comes with mathematical guarantees. They are usually bounds for the approximation errors (linked to the expression power of the hypothesis space) or the generalization error (linked to the robustness of the algorithm with respect to variations of the sample set). These can be also guarantees about the uncertainty around the parameters of the model (represented by confidence intervals for instance). Linear approaches are, in this scope, the most statistically interpretable ML algorithm. Robustness properties of statistical models are also desirable for interpretable models. This is especially the case when considering explanations based on counterfactual examples. Given a binary classifier and an example $e$, the counterfactual of $e$ is the closest example to $e$ with respect to a metric  that is labeled by the classifier with the opposite label of $e$ (the counterfactual is not necessarily in the dataset). Consider for instance a model that determines if a credit is allowed or not with respect to the profile of a customer and a client to which the credit is not granted. The counterfactual in this case answers the question of what is the minimal change on his profile that would ensure that the credit is  granted. If the model is based on propositional logic rules, the counterfactual will correspond to a minimal change of the considered example representation in Boolean logic. In this case, the counterfactual is an understandable explanation for the prediction. In deep-learning, the counterpart of counterfactuals are adversarial examples \cite{goodfellow2014explaining}. In most situations, an adversarial example corresponds to an imperceptible modification of the considered example. From this point of view, the lack of robustness of deep networks makes explanations based on counterfactuals very difficult to obtain.

A decision function learned by a system is assumed to be interpretable if it is simple, often meaning a linear model with few parameters, or if it is expressed with terms or rules that a domain expert is supposed to understand, like in rule-based systems or in decision trees. One influencing work promotes the use of locally linear models in order to offer some interpretability even to globally non linear models \cite{guestrin-kdd16}. 

Explainability can be understood at two levels. Either at the level of the learning algorithm itself that should be easily understandable by ML experts as well as by practitioners or users of the system. Or at the level of the learned model that, for instance, could incorporate causal relationships. One question is: is it possible to extract causal relationships from data alone, without some prior knowledge that suggest those relationships? Judea Pearl \cite{pearl2009causality,pearl2016causal,pearl2018book} argues that this is not possible, but gives to the ML techniques the role of identifying possible correlations between variables in huge data sets that are impossible to sift through for human experts. A recent work \cite{lopez2017discovering} suggests that it would be possible to identify the direction of a causal relationship from observational data. However, the necessary interplay between ML, prior knowledge and reasoning is still a matter of debate.

When dealing with high-dimensional structured data (such as image or text), interpretable and explainable approaches (in a statistical or a KR point of view) are known to be less effective than heavy numerical approaches such as bagging (random forest, gradient boosting) or deep learning \cite{LeCunBH15,Goodfellow-et-al-2016}. Deep learning models are neither explainable nor interpretable due to the large number of parameters and their entanglement. There also exist few statistical results for deep learning and the currently known properties are restricted to very specific architectures (see \cite{EldanS15} for instance).\\

Some approaches have been proposed for improving the explainability of a deep learning algorithms (or in some cases any black box algorithm). A first solution for explaining the solution is to analyze the sensitivity of the prediction with respect to small variations of the input. For instance, activation maps \cite{ZhouKLOT15} will emphasise the most important pixel of a picture for a given prediction. Although this type of representation is easily readable, there are some cases where this is not enough for explanation. Another solution is to approximate the model (globally or locally) with an explainable one. Even when the approximation error is reasonable, we have no guarantee that the interpretation associated with the surrogate model is related to the way the initial model makes the prediction. In \cite{lopez2017discovering}, authors propose to locally replace the model with a surrogate interpretable model. This allows to reduce the approximation error but it is based on a neighbourhood notion that can be difficult to define in high-dimensional structured spaces. Moreover, using an interpretable/explainable model is not always a guarantee for explainable prediction. Indeed, a linear function with millions of parameters or a set of rules with thousand of literals may be not readable at all.
%
%
%
%
%
%
%
%
%
%
%

\section{Some Methodologies Combining Reasoning Principles and Learning}
The idea of combining KRR ingredients with ML tools is not new. It has been done in different ways. This section presents a series of examples of methodologies mixing KRR and ML ideas, with no intent to be exhaustive however. 
Each subsection roughly follows the same structure, stating the 
goal of the methodology, presenting its main aspects, 
and identifying the KR and the ML parts.

\subsection{Injecting Knowledge in Learning} 

Induction cannot be made to work without prior knowledge that restrains the space of models to be explored. Two forms of prior knowledge, aka. biases, are distinguished: \textit{representation biases} that limit the expressiveness of the language used to express the possible hypotheses on the world, and \textit{search biases} that control how the hypothesis space is explored by the learning algorithm. 

\textit{Representation biases} can take various forms. They can directly affect the language in which the possible hypotheses can be expressed. For instance, ``hypotheses can involve a maximum of two disjuncts''. In the same way, but less declaratively, looking for linear models only is a severe representation bias. Often, one does not want to be so strict, and prefers to favour more flexible preference criteria over the space of possible hypotheses. Generally, this is expressed through a regularized optimisation criterion that balances a measure of fit of the model to the data, and a measure of fit of the model to the bias. For instance, the following quality measure over linear hypotheses $h(x) = \beta_0 + \sum_{j=1}^{p} \beta_j x_{i,j}$ for regression favours hypotheses that involve fewer parameters: 

\begin{equation*}
 R(h) \; = \; \underbrace{\frac{1}{2}\sum_{i=1}^{m} \left( y_i - \beta_0 - \sum_{j=1}^{p} \beta_j x_{i,j} \right)^2}_{\text{fit to the data}} \; + \; \underbrace{\lambda \sum_{j=1}^{p} ||\beta_j||_0}_{\text{Favors models with few non zero parameters}} 
\end{equation*}
where the $L_0$ norm $||.||_0$ counts the nonzero parameters $\beta_j$.

The \textit{search bias} dictates how the learning algorithm explores the space of hypotheses. For instance, in the case of neural networks, the search starts with a randomly initialized neural network and then proceeds by a gradient descent optimization scheme. In some other learning methods, such as learning with version space, the search uses generalization relations between hypotheses in order to converge towards good hypotheses. In this latter case, it is more easy to incorporate prior knowledge from the experts. Indeed, the exploration of the hypothesis space is akin to a reasoning process, very much like theorem proving.

%
%
\subsection{Inductive Logic Programming} \label{ILP}

Inductive Logic Programming (ILP) (see \cite{deraedt-jlp94,deraedt08,Dzeroski01} for general presentations) is a subfield of ML that aims at learning models expressed in (subsets of) First Order Logic. It is an illustration of Symbolic Learning, where the hypotheses space is discrete and structured by a generality relation. The aim is then to find a hypothesis that covers the positive examples (it is then said to be complete) and rejects the negative ones (it is said to be consistent). The structure of the hypothesis space allows to generalize an incomplete hypothesis, so as to cover more positive examples, or to specialize an inconsistent hypothesis in order to exclude negative covered examples. The main reasoning mechanism is induction in the sense of generalization (subsumption). 

In ILP, examples and models are represented by clauses. Relying on First Order Logic allows to model complex problems, involving structured objects (for instance to determine whether a molecule is active or not, a system must take into account the fact it is composed of atoms with their own properties and shared relations), or involving objects in relation with each other (a social network or temporal data). Reasoning is a key part of ILP. First, the search for a model is usually performed by exploring a search space structured by a generality relation. A key point is then the definition of a generality relation between clauses. The more natural definition of subsumption should be expressed in terms of logical consequences, which allows comparing the models of both formula, but since the problem is in general not decidable, the notion of $\theta$-subsumption, as introduced in \cite{Plotkin70} is usually preferred: a clause $C_1$ is more general that a clause $C_2$ if there exists a substitution $\theta$ such that $C_1.\theta \subseteq C_2$. In this definition a clause, i.e. a disjunction of literals, is represented by its set of literals. For instance, the rule $par(X,Y), par(Y,Z) \rightarrow grand\_par(X,Z)$ $\theta$-subsumes $par(john,ann), par(ann,peter), par(john, luc) \rightarrow grand\_par(john,peter)$.Indeed, the first one leads to the clause $\neg par(X,Y) \vee \neg par(Y,Z) \vee grand\_par(X,Z)$ and the second one $\neg par(john,ann) \vee \neg par(ann,peter) \vee \neg par(john, luc) \vee grand\_par(john,peter)$.  Second, expert knowledge  
can be expressed using facts (ground atoms) or by rules, or yet reasoning mechanisms to be applied. This can be illustrated by the well-known systems FOIL \cite{Quinlan1996} and Progol \cite{Muggleton95}

ILP, and more generally Symbolic Learning, has thus some interesting properties. First, the model is expressed in logic and therefore is claimed to be easily understandable by a user (See for instance \cite{MuggletonSZTB18} for an interesting study of the comprehensibility or not of programs learned with ILP). 
Second, expert knowledge can be easily expressed by means of clauses and integrated into the learning algorithm.
Although initially developed for the induction of logic programs, it has now shown its interest for learning with structured data. 

However, ILP suffers from two drawbacks: the complexity of its algorithms and its inability to deal with uncertain data. Several mechanisms have been introduced to reduce the complexity, as for instance the introduction of syntactic biases, restricting the class of clauses that can be learned. Another interesting idea is propositionnalization, introduced in \cite{LavracDG91} and then developed for instance in the system RSD \cite{ZeleznyL06}. It is a process that transforms a relational problem into a classical attribute-value problem by the introduction of new features capturing relations between objects. Once the transformation performed, any supervised learner can be applied to the problem. The main difficulty is then to define these new features. 

 This last problem has led to the emergence of Statistical Relational Learning \cite{GetoorTaskar:book07,Raedt2008} that aims at coupling ILP with probabilistic models. Many systems have been developed, extending naive Bayesian classifier \cite{LachicheF02}, Bayesian Networks \cite{FierensBBR05} or Markov Logic Networks \cite{RichardsonD06} or developing new probabilistic framework as in Problog \cite{RaedtKT07}. In all these works, inference and learning are tightly connected since learning parameters requires to maximize the likelihood for generative learning (estimation of the probabilities to generate the data, given a set of parameters),
 or the conditional likelihood in case of discriminative learning (estimation of the probabilities of the labels given the data).
 Optimizing the parameters thus require at each step to estimate the corresponding probabilities. This has led to intensive research on the complexity of inference.


In the last decade new works have emerged linking deep learning and Inductive Logic Programming. Two directions are investigated. The first one relies on propositionnalization as in \cite{KaurKKKCN17}: first, a set of interpretable rules is built through a Path Ranking Algorithm and then the examples are transformed into an attribute-value representation. Two settings are considered: an existential one making the feature true if an instantiation of this path exists in the example, a counting one that counts the number of times the random walk is satisfied. Once this transformation is performed, a multilayered discriminative RBM can be applied. The second direction, illustrated in \cite{SourekSZSK17}, consists in encoding directly facts and ground rules as neurons. Aggregation neurons allow combining rules with the same head and the activation function are approximations of \L ukasiewicz fuzzy logic. 


\subsection{Neuro-Symbolic Reasoning}\label{neurosy}

Several works have proposed to combine learning and reasoning by studying schemes to translate logical representations of knowledge into neural networks. A long-term goal of a series of works on neural-symbolic integration, surveyed for instance by \cite{garcez-corr17}, is ``to provide a coherent, unifying view for logic and connectionism \ldots \textit{[in order to]} \ldots produce better computational tools for integrated ML and reasoning." Typical works propose translation algorithms from a symbolic to a connectionist representation, enabling the use of computation methods associated with neural networks to perform tasks associated with the symbolic representation.

Early works in this vein are \cite{Pinkas:neur-comp91,Pinkas:art-int95,BornscheuerHolldoblerKalinkeStrohmaier:aut-ded-basis-appII98,HolldoblerKalinkeStorr:app-int99}. Bornscheuer {\em et al.}  \cite{BornscheuerHolldoblerKalinkeStrohmaier:aut-ded-basis-appII98} show for example how an instance of the Boolean satisfiability problem can be translated into a feed-forward network that parallelizes GSAT, a local-search algorithm for Boolean satisfiability. They also show that a normal logic program $ P $ can be turned into a connectionnist network that can approximate arbitrarily well the semantics of the program.

The paper \cite{dAvilaGarcezZaverucha:app-int99} exploits this idea to represent propositional logic programs with recurrent neural networks (RNNs) which can be used to compute the semantics of the program. They show that this program can also be used as background knowledge to learn from examples, using back-propagation. Essentially, the RNN defined to represent a logic program $ P $ has all atoms of $ P $ in the input layer; one neuron, a kind of ``and'' gate, for each rule in a single hidden layer; and one neuron for every atom in the output layer, these neurons working like ``or'' gates. Re-entrant connections from an atom in the output layer to its counterpart in the input layer enable the  chaining of rules.

Franca {\em et al.} \cite{FrancaZaveruchadAvilaGarcez:mach-learn14} extend these results to first-order programs, using a propositionalization method called Bottom Clause Propositionalization. In \cite{dAvilaGarcezLambGabbay:tcs07,dAvilaGarcezLamb:nips03}, methods are proposed to translate formulas that contain modalities into neural networks, enabling the representation of time and knowledge, and the authors in \cite{dAvilaGarcezLambGabbay:his03,dAvilaGarcezLambGabbay:tcs06} show that there exists a neural network ensemble that computes a fixed-point semantics of an intuitionistic theory. Pinkas and Cohen \cite{PinkasCohen:flap19} perform experiments with so-called higher-order sigma-pi units (which compute a sum of products of their inputs) instead of hidden layers, for planning problems on simple block-world problems: the number of units is fixed at the design time, and is a function of the maximum number of blocks and the maximum number of time steps; for example, for every pair $ (b_1,b_2) $ of possible blocks and every time step $ t $, there is a unit representing the proposition $ \text{above}(b_1,b_2,t) $; their results indicate that a learning phase enables the network to approximately learn the constraints with a reasonable amount of iterations, which speeds up the computation of an approximate solution for subsequent instances.

A new direction of research has recently emerged, bringing in particular to Statistical Relational Learning \cite{GetoorTaskar:book07,deraedt-bk16} the power of tensor networks. For instance, in \cite{cohen-nips17,lukasiewicz-corr17}, they use recursive tensor networks to predict classes and / or binary relations from a given knowledge base. It has also been proposed in, e.g., \cite{DiligentiGoriMagginiRigutini:mach-learn12,SocherChenManningNg:nips13,SerafiniAvilaGarcez:aiia16} to depart from the usual semantics of logic based on Boolean truth values in order to use the numerical operators of neural networks. The universe of an interpretation of first-order logic can be a set of vectors of real numbers, and the truth values of predicates can be real numbers in the interval $ [0,1] $; truth values of general formulas can then be defined using usual operators of fuzzy logic. Donadello {\em et al.} \cite{DonadelloSerafinidAvilaGarcez:ijcai17} describe how this approach can be used to learn semantic image interpretation using background knowledge in the form of simple first-order formulas.


\subsection{Formal Concept Analysis}

Formal Concept Analysis (FCA) \cite{GWil,FKHKN}  is another example of a setting that stands in between KRR and ML concerns. It offers a mathematical framework that is based on the duality between a set of objects or items and a set of descriptors. In the basic setting, we start from a formal context which is a relation linking objects and (Boolean) attributes (or properties). Thus a formal context constitutes a simple repository of data. A concept is formalized as a pair composed of a set of attributes and a set of objects representing the intention and the extension of the concept respectively; it has the property that these objects and only them satisfy the set of attributes and this set of attributes refers to these objects and only them. Such a set of attributes is called a closed pattern or a closed itemset. More precisely, two operators, forming a Galois connection, respectively associate their common descriptors to a subset of objects, and the set of objects that satisfy all of them to a subset of descriptors. 
 In an equivalent way, a pair made of a set of objects and a set of attributes is a formal concept if and only if their Cartesian product forms a maximal rectangle for set inclusion in the formal context. The set of concepts forms a complete lattice.
 
 Formal Concept Analysis is deeply rooted in Artificial Intelligence by the formalization of the notion of concept. Recent years have witnessed a renewed interest in FCA with the emergence of Pattern Mining: it has been shown that the set of closed itemsets forms a condensed representation of the set of itemsets, thus reducing the memory space for storing them. Moreover it a possible to define an equivalence relation between itemsets (two itemsets are equivalent if they share the same closure) and from the equivalence classes, it becomes possible to extract all the exact association rules (rules with a confidence equal to 1). See \cite{GuiguesDuquenne,PasquierBTL99,BastidePTSL00}
 
 Two extensions are especially worth mentioning. One uses fuzzy contexts, where the links between objects and attributes are a matter of degree \cite{Beloh}. This may be useful for handling numerical attributes \cite{MessaiDNS08}. Another extension allows for structured or logical descriptors using so-called pattern structures \cite{GanterK01,FerreR04,AssaghirKP10}. Besides, operators other than the ones defining formal concepts make sense in formal concept analysis, for instance to characterize independent subcontexts \cite{DuboisP12}.


\subsection{Rule-Based Models }
Knowledge representation  by if-then rules is a  format whose importance was early acknowledged in the history of AI, with the advent of rule-based expert systems. Their modelling has raised the question of the adequacy of classical logic for representing them, especially in case of uncertainty where conditioning is often preferred to material implication. Moreover, the need for rules tolerating exceptions, or expressing gradedness, such as default rules and fuzzy rules has led KRR to develop tools beyond classical logic. 

\subsubsection{Default rules} Reasoning in a proper way with default rules (i.e., having potential exceptions) was a challenging task for AI during three decades \cite{BMT2011}. Then a natural question is: can rules having exceptions extracted from data be processed by a nonmonotonic inference system yielding new default rules? How can we insure that these new rules are still agreeing with the data? The problem is then to extract genuine default rules that hold 
in a Boolean database. It does not just amount to mining association rules with a sufficiently high confidence level. 
We have to guarantee that any new default rule that is 
 deducible from the set of extracted default rules is indeed valid with respect to the database.
 To this end, we need a probabilistic semantics for nonmonotonic inference. It has been shown \cite{BenferhatDP99} that 
default rules of the form ``if $p$ then generally $q$'', denoted by $p \snake q$, 
where $\snake$ obey the postulates of preferential inference \cite{KLM90},  
have both 
\begin{enumerate}
    \item a possibilistic semantics expressed by the constraint $\Pi(p \wedge q) > \Pi(p \wedge \neg q)$, for any max-decomposable possibility measure $\Pi$ ($\Pi(p \vee q) = \max(\Pi(p), \Pi( q))$),
    \item a probabilistic semantics expressed by the constraint $Prob(p \wedge q) > Prob(p \wedge \neg q)$ for any \emph{big-stepped probability} $Prob$.
\end{enumerate}  This is a very special kind of probability such that if
$p_1> p_2 > ... > p_{n-1} \geq p_n$ (where $p_i$ is the probability of one of the $n$ possible worlds), the following inequalities hold 
$\forall i =1, n-1, \ p_i > \Sigma_{j=i,n} \ p_j$. Then, one can safely infer a new default $p \snake q$ from a set of defaults 
 $\Delta=\{p_k \snake q_k | k=1, K\}$ if and only if the constraints modeling $\Delta$ entail the constraints modeling $p \snake q$.
Thus, extracting defaults amounts to looking for big-stepped probabilities, by clustering lines describing items in Boolean tables, so as to find default rules, see \cite{BenferhatDLP03} for details. Then the rules discovered are genuine default rules
that can be reused in a nonmonotonic inference system, and can be encoded in possibilistic logic (assuming rational monotony for the inference relation).

It may be also beneficial to rank-order a set of rules expressed in the setting of classical logic in order to handle exceptions in agreement with nonmonotonic reasoning. This what has been proposed in \cite{SerrurierP07} where a new formalization of inductive logic programming (ILP) in first-order possibilistic logic allows us to handle exceptions by means of prioritized rules. Indeed, in classical first-order logic, exceptions of the rules 
can be assigned to more than one class, even if only one is the right one, which is not correct. The possibilistic formalization provides a sound encoding of non-monotonic reasoning that copes with rules with exceptions and prevents an example from being classified in more than one class.

Possibilistic logic \cite{DLP94} is also a basic logic for handling epistemic uncertainty. It has been established that any set of Markov logic formulas \cite{RichardsonD06}
can be exactly translated into possibilistic logic formulas 
\cite{KuzelkaDS15,DuboisPS17}, thus providing an interesting bridge between KRR and ML concerns.

\subsubsection{Fuzzy rules}

The idea of fuzzy if-then rules was first proposed by L. A. Zadeh \cite{Zadeh73}. They are rules whose conditions and /or conclusions express fuzzy restrictions on the possible values of variables. Reasoning with fuzzy rules is based on a combination / projection mechanism \cite{Zadeh1979} where the fuzzy pieces of information (rules, facts) are conjunctively combined and projected on variables of interest. Special views of fuzzy rules have been used when designing fuzzy rule-based controllers: fuzzy rules may specify the fuzzy graph of a control law which once applied to an input yields a fuzzy output that is usually defuzzified 
\cite{MamdaniA1975}. Or rules may have precise conclusions that are combined on the basis of the degrees of matching between the current situation and the fuzzy condition parts of the rules \cite{TakagiSug85}. In both cases, an interpolation mechanism is at work, implicitly or explicitly \cite{Zadeh92}. Fuzzy rules-based controllers are universal approximators \cite{Castro1995}. The functional equivalence between a radial basis function-based neural network and a fuzzy inference system has been established under certain conditions \cite{JangSun93}. 

Moreover, fuzzy rules may provide a rule-based interpretation \cite{dAlAndNad94,HuangaXing2002} for (simple) neural nets, and neural networks can be used for extracting fuzzy rules from the training data \cite{Pedrycz98,LengMP2005}.
Regarding neural nets, let us also mention a non-monotonic inference view \cite{BalkeniusG91,Gardenfors91}.

Association rules \cite{AgrawalEtal93,HaHa78} describe relations between variables together with confidence and support degrees. See \cite{DuboisHP06} for the proper assessment of confidence and support degrees in the fuzzy case. In the same spirit, learning methods for fuzzy decision trees have been devised in   \cite{MarsalaB10}, in the case of numerical attributes. The use of fuzzy sets to describe associations between data and decision trees may have some interest: extending the types of relations that may be represented, making easier the interpretation of rules in linguistic terms \cite{DuboisPS05}, and avoiding unnatural boundaries in the partitioning of the attribute domains.

There are other kinds of fuzzy rules whose primary goal is not to approximate functions nor to quantify associations, but rather to offer representation formats of interest. This is, for instance, the case of gradual rules, which express statements of the form ``the more $x$ is $A$, the more $y$ is $B$, where $A$, and $B$ are gradual properties modelled by fuzzy sets \cite{SerrurierDPS07,NinLP10}. 

\subsubsection{Threshold rules}
Another format of interest is the one of multiple threshold rules, i.e., selection rules of the form ``if $x_1\geq \alpha_1$ and $\cdots$ $x_j\geq \alpha_j$ and $\cdots$ then $y\geq \gamma$'' (or deletion rules of the form `if $x_1\leq \beta_1$ and $\cdots$ $x_j\leq \beta_j$ and $\cdots$ then $y\geq \delta$''), which are useful in monotone classification / regression problems \cite{GrecoIS06,vcdomlem}. Indeed when dealing with data that are made of a collection of pairs $(x^k, y_k), k = 1, ..., N$, where $x^k$ is a tuple $(x_1^k, ..., x_n^k)$ of feature evaluations of item $k$, and where $y$ is assumed to increase with the $x_i$'s in the broad sense, it is of interest of describing the data with such rules of various lengths. 
It has been noticed \cite{GrecoMS04,DuboisPR14} that, once the numerical data are normalized between 0 and 1, rules where all (non trivial) thresholds are equal can be represented by Sugeno integrals (a generalization of weighted min and weighted max, which is a qualitative counterpart of Choquet integrals \cite{GL10}). Moreover, it has been shown recently \cite{BrabantCDPR18} that generalized forms of Sugeno integrals are able to describe a global (increasing) function, taking values on a finite linearly ordered scale, under the form of general thresholded rules. Another approach, in the spirit of the version space approach \cite{Mitchell}, provides a bracketing of an increasing function by means of a pair of Sugeno integrals \cite{PradeRSR09,PradeRS09}.

\subsection{Uncertainty in ML: in the data or in the model}



We will focus on two aspects in which cross-fertilisation of ML with KRR could be envisaged: uncertainty in the data and uncertainty in the models/predictions. 

\subsubsection{Learning under uncertain and coarse data}
In general, learning methods assume the data to be complete, typically in the form of examples 
being precise values (in the unsupervised case) or precise input/output pairs (in the supervised case). There are however various situations where data can be expected to be uncertain, such as when they are provided by human annotators in classification or measured by low-quality sensors, or even missing, such as when sensors have failed or when only a few examples could be labelled. An important remark is that the uncertainty attached to a particular piece of data can hardly be said to be of objective nature (representing frequency) as it has a unique value, and this even if this uncertainty is due to an aleatory process. 

While the case of missing (fully imprecise) data is rather well-explored in the statistical~\cite{little2019statistical} and learning~\cite{chapelle2006semi} literature, the general case of uncertain data, where this uncertainty can be modelled using different representation tools of the literature, largely remains to be explored. In general, we can distinguish between two strategies:
\begin{itemize}
 \item The first one intends to extend the precise methods so that they can handle uncertain data, still retrieving a precise model from them. The most notable approaches consist in either extending the likelihood principle to uncertain data (e.g., \cite{denoeux2013maximum} for evidential data, or \cite{couso2018general} for coarse data), or to provide a precise loss function defined over partial data and then using it to estimate the empirical risk, see for instance \cite{hullermeier2014learning,cour2011learning,cid2012proper}. Such approaches are sometimes based on specific assumptions, usually hard to check, about the process that makes data uncertain or partial. Some other approaches such as the evidential likelihood approach outlined in \cite{denoeux2013maximum} do not start from such assumptions, and simply propose a generic way to deal with uncertain data. We can also mention transductive methods such as the evidential $K$-nearest neighbour ($K$-NN) rule \cite{denoeux95a,denoeux01a,denoeux19f}, which allows one to handle partial (or ``soft'') class labels without having to learn a model.
 \item The second approach, much less explored, intends to make no assumptions at all about the underlying process making the data uncertain, and considers building the set of all possible models consistent with the data. Again, we can find proposals that extend probability-based approaches~\cite{de2004updating}, as well as loss-based ones~\cite{couso2016machine}. The main criticism one could address to such approaches is that they are computationally very challenging. Moreover they do not yield a single predictive model, making the prediction step potentially difficult and ill-defined, but also more robust. 
\end{itemize}
The problem of handling partial and uncertain data is certainly widely recognised in the different fields of artificial intelligence, be it KRR or ML. One remark is that mainstream ML has, so far, almost exclusively focused on providing computationally efficient learning procedures adapted to imprecise data given in the form of sets, as well as the associated assumptions under which such a learning procedure may work~\cite{liu2014learnability}. While there are proposals around that envisage the handling of more complex form of uncertain data than just sets, such approaches remain marginal, at least for two major reasons:
\begin{itemize}
 \item More complex uncertainty models require more efforts at the data collection step, and the benefits of such an approach (compared to set-valued data or noisy precise data) do not always justify the additional efforts. However, there are applications in which the modeling of data uncertainty in the belief function framework does improve the performances in classification tasks \cite{cherfi12,quost17}. Another possibility could be that those data are themselves prediction of an uncertain model, then used in further learning procedures (such as in stacking~\cite{dvzeroski2004combining}); 
 \item Using more complex representations may involve a higher computational cost, and the potential gain of using such representations is not always worth the try. However, some specific cases approaches such as the EM algorithm \cite{denoeux2013maximum} or the $K$-NN rule \cite{denoeux19f} in the evidential setting, make it possible to handle uncertain data without additional cost.
\end{itemize}
\subsubsection{Uncertainty in the prediction model}
Another step of the learning process where uncertainty can play an important role is in the characterisation of the model or its output values. In the following, we will limit ourselves to the supervised setting where we search to learn a (predictive) function $f:\mathcal{X} \to \mathcal{Y}$ linking an input observation $x \in \mathcal{X}$ to an output (prediction) $y \in \mathcal{Y}$. Assessing the confidence one has in a prediction can be important in sensitive applications. This can be done in different ways:
\begin{itemize}
 \item By directly impacting the model $f$ itself, for instance associating to every instance $x$ not a deterministic prediction $f(x)$, but an uncertainty model over the domain $\mathcal{Y}$. The most common one is of course probabilities, but other solutions such as possibility distributions, belief functions or convex sets of probabilities are possible;
 \item By allowing the prediction to become imprecise, the main idea behind such a strategy being to have weaker yet more reliable predictions. In the classical setting, this is usually done by an adequate replacement of the loss function~\cite{ha1997optimum,grandvalet2009support}, yet recent approaches take a different road. For instance, imprecise probabilistic approaches consider sets of models combined with a skeptic inference (also a typical approach in KR), where a prediction is rejected if it is so for every possible model~\cite{corani2012bayesian}. Conformal prediction~\cite{shafer2008tutorial} is another approach that can be plugged to any model output to obtain set-valued predictions. Approaches to quantify statistical predictions in the belief function framework are described in \cite{kanjana16,xu16,denoeux18b}.
\end{itemize}
If such approaches are relatively well characterised for the simpler cases of multi-class classification, extending them to more complex settings such as multi-label or ranking learning problems that involve combinatorial spaces remain largely unexplored, with only a few contributions~\cite{cheng2012label,antonucci2017multilabel}. It is quite possible that classical AI tools such as SAT or CSP solvers could help to deal with such combinatorial spaces. 

\subsection{Case-Based Reasoning, Analogical Reasoning and Transfer Learning}
Case-based reasoning (CBR for short), e.g., \cite{AamodtP94} is a form of reasoning that exploits data (rather than knowledge) under the form of cases, often viewed as pairs $\langle$problem, solution$\rangle$. When one seeks for potential solution(s) to a new problem, one looks for previous solutions to similar problems in the repertory of cases, and then adapts them (if necessary) to the new problem. 

Case-based reasoning, especially when similarity is a matter of degree, thus appears to be close to k-NN methods and instance-based learning \cite{denoeux95a,HullermeierDP02}. 
 The k-NN method is a prototypical example of transduction, i.e., the class
of a new piece of data is predicted on the basis of previously observed data, without any attempt at inducing a generic model for the observed data. The term transduction was coined in \cite{Gammerman98learningby}, but the idea dates back to Bertrand Russell \cite{Russell1912}. 

Another example of transduction is analogical proportion-based learning. Analogical proportions are statements of the form ``$a$ is to $b$ as $c$ is to $d$'', 
often denoted by $a:b::c:d$, 
 which express that 
 ``$a$ differs from $b$ as $c$ differs from $d$ and $b$ differs from $a$ as $d$ differs from $c$''. This statement can be encoded into a Boolean logical expression \cite{MicPraECSQARU2009,PraRicLU2013} 
which is true only for the 6 following assignments $(0, 0, 0, 0)$, $(1, 1, 1, 1)$, $(1, 0, 1, 0)$, $(0, 1, 0, 1)$, $(1, 1, 0, 0)$, and $(0, 0, 1, 1)$ for $(a, b, c, d)$. 
Note that they are also compatible with the arithmetic proportion definition $a - b =c - d$, where $a - b \in \{-1, 0,1\}$, which is not a Boolean expression.
Boolean Analogical proportions straightforwardly extend to vectors of attributes values such as $\vec{a}=(a_1, ..., a_n)$, by stating 
$\vec{a}:\vec{b}::\vec{c}:\vec{d} \mbox{ iff } \forall i \!\in\! [1,n], ~ a_i:b_i::c_i:d_i$. The basic analogical inference pattern \cite{StrYvoReport2005}, is then
$$\frac{\forall i \in \{1,..., p\}, ~~a_i : b_i :: c_i : d_i \mbox{ holds}}{\forall j \in \{p+1,..., n\}, ~~a_j : b_j :: c_j : d_j \mbox{ holds}}$$
Thus analogical reasoning amounts to finding completely informed triples $(\vec{a}, \vec{b}, \vec{c})$ appropriate for inferring the missing value(s) in $\vec{d}$. When there exist several suitable triples, possibly leading to distinct conclusions, one may use a majority vote for concluding. This inference method extends to analogical proportions between numerical values, and the analogical proportion becomes graded \cite{DuboisPR16}. It has been successfully applied, for Boolean, nominal or numerical attributes, to classification \cite{MicBayDel2008,BouPraRicIJAR2017b} 
(then the class $cl(\vec{x})$ (viewed as a nominal attribute) is the unique solution, when it exists, such as $cl(\vec{a}):cl(\vec{b})::cl(\vec{c}):cl(\vec{x})$ holds), and more recently to case-based reasoning \cite{LieberNPR18} and to preference learning \cite{FH18,BounhasPP18}.
It has been theoretically established that analogical classifiers \emph{always} yield exact prediction for Boolean affine functions (which includes x-or functions), and only for them \cite{CouHugPraRicIJCAI2017}. Good results can still be obtained in other cases \cite{CouHugPraRicIJCAI2018}.
Moreover, analogical inequalities \cite{PraRicIJAR2018} of the form ``$a$ is to $b$ at least as much as $c$ is to $d$'' might be useful for describing relations between features in images, as in \cite{LawTC17}.

\smallskip
The idea of transfer learning, which may be viewed as a kind of analogical reasoning performed at the meta level, is to take advantage of what has been learnt on a source domain in order to improve the learning process in a target domain related to the source domain.
When studying a new problem or a new domain, it is natural to try to identify a related, better mastered, problem or domain from which, hopefully, some useful information can be called upon for help. The emerging area of transfer learning is concerned with finding methods to transfer useful knowledge from a known source domain to a less known target domain. 

The easiest and most studied problem is encountered in supervised learning. There, it is supposed that a decision function has been learned in the source domain and that a limited amount of training data is available in the target domain. For instance, suppose we have learnt a decision function that is able to recognize poppy fields in satellite images. Then the question is: could we use this in order to learn to recognize cancerous cells in biopsies rather than to start anew on this problem or when few labeled data is available in the biological domain?

This type of transfer problem has witnessed a spectacular rise of interest in recent years thanks both to the big data area that makes lots of data available in some domains, and to the onset of deep neural networks. In deep neural networks, the first layers of neuron like elements elaborate on the raw input descriptions by selecting relevant descriptors, while the last layers learn a decision function using these descriptors. Nowadays, most transfer learning methods rely on the idea of transferring the first layers when learning a new neural network on the target training data, adjusting only the last layers. The underlying motivation is that the descriptors are useful in both the source and target domains and what is specific is the decision function built upon these descriptors. But it could be defended just as well that the decision function is what is essential in both domains while the ML part should concentrate on learning an appropriate representation. This has been achieved with success in various tasks \cite{cornuejols2018a}. 

One central question is how to control what should be transferred. A common assumption is that transfer learning should involve a minimal amount of change of the source domain knowledge in order for it to be used in the target domain. Several ways of measuring this ``amount of change'' have been put forward (see for instance \cite{courty2016optimal,shen2017wasserstein}), but much work remains to be done before a satisfying theory is obtained.

One interesting line of work is related to the study of causality. Judea Pearl uses the term ``transportability'' instead of transfer learning, but the fundamental issues are the same. Together with colleagues, they have proposed ways of knowing if and what could be transferred from one domain to another \cite{pearl2018book}. The principles rely on descriptions of the domains using causal diagrams. Thanks to the ``do-calculus'', formal rules can be used in order to identify what can be used from a source domain to help solve questions in the target domain. 
One foremost assumption is that causality relationships capture deep knowledge about domains and are somewhat preserved between different situations. For instance, proverbs in natural language are a way of encapsulating such deep causality relationships and their attractiveness comes from their usefulness in many domains or situations, when properly translated.

\section{Examples of KRR/ML Synergies}
In the previous section, we have surveyed various paradigms where KRR and ML aspects are intricately entwined together. In this section, we rather review examples of hybridizations where KRR and ML tools cooperate. In each case, we try to identify the purpose, the way the KRR and ML parts interact, and the expected benefits of this synergy.

\subsection{Dempster-Shafer Reasoning and Generalized Logistic Regression Classifiers}


The theory of belief functions originates from Dempster's seminal work \cite{dempster67a} who proposed, at the end of the 1960’s, a method of statistical inference that extends both Fisher’s fiducial inference and Bayesian inference. 
In a landmark book, Shafer \cite{shafer76} developed Dempster’s mathematical framework and extended its domain of application by showing that it could be proposed as a general language to express “probability judgements” (or degrees of belief) induced by items of evidence. This new theory rapidly became popular in Artificial Intelligence where it was named “Dempster-Shafer (DS) theory”, evidence theory, or the theory of belief functions. DS theory can be considered from different perspectives:
\begin{itemize}
\item	A belief function can be defined axiomatically as a Choquet monotone capacity of infinite order \cite{shafer76}. 
\item	Belief functions are intimately related to the theory of random sets: any random set induces a belief function and, conversely, any belief function can be seen as being induced by a random set \cite{nguyen78}. 
\item Sets are in one-to-one correspondence with so-called ``logical'' belief functions, and probability measures are special belief functions. 
A belief function can thus be seen both as a generalised probability measure and as a generalised set; it makes it possible to combine reasoning mechanisms from probability theory (conditioning, marginalisation), with set-theoretic operations (intersection, union, cylindrical extension, interval computations, etc.)
\end{itemize}

DS theory thus provides a very general framework allowing us to reason with imprecise and uncertain information. In particular, it makes it possible to represent states of knowledge close to total ignorance and, consequently, to model situations in which the available knowledge is too limited to be properly represented in the probabilistic formalism. Dempster's rule of combination \cite{shafer76} is an important building block of DS theory, in that it provides a general mechanism for combining independent pieces of evidence.

The first applications of DS theory to machine learning date back to the 1990's and concerned classifier combination \cite{xu92,rogova94}, each classifier being considered as a piece of evidence and combined by Dempster's rule (see, e.g., \cite{quost11} for a refinement of this idea taking into account the dependence between classifier outputs). In \cite{denoeux95a}, Den{\oe}ux combined Shafer's idea of evidence combination with distance-based classification to introduce the evidential $K$-NN classifier \cite{denoeux95a}. In this method, each neighbour of an instance to be classified is considered as a piece of evidence about the class of that instance and is represented by a belief function. The $K$ belief functions induced by the $K$ nearest neighbour are then combined by Dempster's rule. Extensions of this simple scheme were later introduced in \cite{zouhal97,lian16,denoeux19f}.

The evidential $K$-NN rule is, thus, the first example of an ``evidential classifier''. Typically, an evidential classifier breaks down the evidence of each input feature vector into elementary mass functions and combines them by Dempster's rule. The combined mass function can then be used for decision-making. Thanks to the generality and expressiveness of the belief function formalism, evidential classifiers provide more informative outputs than those of conventional classifiers. This expressiveness can be exploited, in particular, for uncertainty quantification, novelty detection and information fusion in decision-aid or fully automatic decision systems \cite{denoeux19f}.

In \cite{denoeux19f}, it is shown that that not only distance-based classifiers such as the evidential $K$-NN rule, but also a broad class of supervised machine learning algorithms, can be seen as evidential classifiers. This class contains logistic regression and its non linear generalizations, including multilayer feedforward neural networks, generalized additive models, support vector machines and, more generally, all classifiers based on linear combinations of input or higher-order features and their transformation through the logistic or softmax transfer function. Such \emph{generalized logistic regression classifiers} can be seen as combining elementary pieces of evidence supporting each class or its complement using Dempster's rule. The output class probabilities are then normalized plausibilities according to some underlying belief function. This ``hidden'' belief function provides a more informative description of the classifier output than the class probabilities, and can be used for decision-making. Also, the individual belief functions computed by each of the features provide insight into the internal operation of classifier and can help to interpret its decisions. This finding opens a new perspective for the study and practical application of a wide range of machine learning algorithms.

\subsection{Maximum Likelihood Under Coarse Data}
When data is missing or just imprecise (one then speaks of \emph{coarse data}), statistical methods need to be adapted. In particular the question is whether one wishes to model the observed phenomenon {\em along with} the limited precision of observations, or {\em despite} imprecision. The latter view comes down to complete the data in some way (using imputation methods). A well-known method that does it is the EM algorithm \cite{Dempster77}. This technique makes strong assumptions on the measurement process so as to relate the distribution ruling the underlying phenomenon and the one ruling the imprecise outcomes.It possesses variants based on belief functions \cite{denoeux2013maximum}. EM is extensively used for clustering (using Gaussian mixtures) and learning Bayesian nets.

However the obtained result, where by virtue of the algorithm, data has become complete and precise, is not easy to interpret. If we want to be faithful to the data and its imperfections, one way is to build a model that accounts for the imprecision of observations, i.e., a set-valued model. This is the case if a belief function is obtained via maximum likelihood on imprecise observations: one optimises the {\em visible likelihood function} \cite{couso2018general}. The idea is to cover all precise models that could have been derived, had the data been precise. Imprecise models are useful to lay bare ignorance when it is present, so as to urge finding more data, but it may be problematic for decision or prediction problems, when we have to act or select a value despite ignorance.

Ideally we should optimize the likelihood function based on the actual values hidden behind the imprecise observations. But such a likelihood function 
is ill-known in the case of coarse data \cite{couso2018general}. In that case, we are bound 
 \begin{itemize}
 \item To make assumptions on the measurement process so as to create a tight link between the hidden likelihood function pertaining to the outcomes of the real phenomenon, and the visible likelihood of the imprecise observations (for instance the CAR (coarsening at random) assumption \cite{HR91}, or the superset assumption  \cite{HullermeierC15}. In that case, the coarseness of the data can be in some sense ignored. See \cite{Jaeger05} for a general discussion.
 \item Or to pick a suitable hidden likelihood function among the ones compatible with the imprecise data, for instance using an optimistic maximax approach that considers that the true sample is the best possible sample in terms of likelihood compatible with the imprecise observation \cite{hullermeier2014learning}. This approach chooses a compatible probability distribution with the minimum of entropy, hence tends to disambiguate the data. On the contrary maximin approach considers that the true sample is the worst compatible sample in terms of likelihood. This approach chooses a compatible probability distribution with the maximum of entropy. Those two approaches suppose extreme point of view on the entropy of the probability distribution. More recently, an approach base on the likelihood ratio that maximize the minimal possible ratio over the compatible probability distribution is proposed in \cite{GuillaumeD18}. This approach achieving a trade-off between these two more extreme approaches and is able to quantify the quality of the chosen probability distribution in regards to all possible probability distribution. In these approaches, the measurement process is ignored.
 \end{itemize}
 See \cite{CousoDH17,HullermeierDC19} for more discussions about such methods for statistical inference with poor quality data.\\
 
 	Besides, another line of work for taking into account the scarcity of data in ML is to use 
	a new cumulative entropy-like function that together considers the entropy of the probability distribution and the uncertainty pertaining to the estimation of its parameters. It 
	 	takes advantage of the ability of a possibility distribution to upper bound a family of probabilities previously estimated from a limited set of examples 
		\cite{SerrurierP13,SerrurierP15}. Such a function takes advantage of the ability of a possibility distribution to upper bound a family of probabilities previously estimated from a limited set of examples and of the link between possibilistic specificity order and entropy \cite{DuboisH07}. This approach enables the expansion of decision trees to be limited when the number of examples at the current final nodes is too small.
\subsection{EM Algorithm and Revision}
Injecting concepts from KRR, when explaining the EM algorithm may help better figure out what it does. In the most usual case, the coarse data are elements of a partition of the domain of values of some hidden variable. Given a class of parametric statistical models, the idea is to iteratively construct a precise model that fits the data as much as possible, by first generating at each step a precise observation sample in agreement with the incomplete data, followed by the computation of a new model obtained by applying the maximum likelihood method to the last precise sample. These two steps are repeated until convergence to a model is achieved.

In \cite{CousoD16}, it has been shown that the observation sample implicitly built at each step can be represented by a probability distribution on the domain of the hidden variable that is in agreement with the observed frequencies of the coarse data. It is obtained by applying, at each step of the procedure, the oldest (probabilistic) revision rule well-known in AI, namely Jeffrey's rule \cite{Jef83}, to the current best parametric model. This form of revision considers a prior probability $p(x, \theta)$ on the domain $\mathcal{X}$ of a variable $X$, and new information made of a probability distribution over a partition $\{A_1, A_2, \dots, A_n\}$ of $\mathcal{X}$ (representing the coarse data). If $p'_i$ is the ``new'' probability of $A_i$, the old distribution $p(x, \theta)$ is revised so as to be in agreement with the new information. The revised probability function is of the form $p'(x, \theta) = \sum_{i = 1}^n p'_i\cdot p (x, \theta|A_i)$. The revision step minimally changes the prior probability function in the sense of Kullback-Leibler relative entropy. 

In the case of the EM algorithm, $p'_i$ is the frequency of the coarse observation $A_i$, and $p(x, \theta)$ is the current best parametric model. The distribution $p'(x, \theta)$ corresponds to a new sample of $X$ in agreement with the coarse observation. In other words the EM algorithm in turn revises the parametric model to make it consistent with the coarse data, and applies maximum entropy to the new obtained sample, thus minimizing the relative (entropic) distance between a parametric model and a probability distribution in agreement with the coarse data.

\subsection{Conceptual Spaces and the Semantic Description of Vector Representations}

Neural networks, and many other approaches in machine learning, crucially rely on vector representations. Compared to symbolic representations, using vectors has many advantages (e.g., their continuous nature often makes optimizing loss functions easier). At the same time, however, vector representations tend to be difficult to interpret, which makes the models that rely on them hard to explain as well. Since this is clearly problematic in many application contexts, there has been an increasing interest in techniques for linking vector spaces to symbolic representations. The main underlying principles go back to the theory of \emph{conceptual spaces} \cite{Gardenfors:conceptualSpaces}, which was proposed by G\"ardenfors as an intermediate representation level between vector space representations and symbolic representations. Conceptual spaces are essentially vector space models, as each object from the domain of discourse is represented as a vector, but they differ in two crucial ways. First, the dimensions of a conceptual space usually correspond to interpretable salient features. Second, (natural) properties and concepts are explicitly modelled as (convex) regions. Given a conceptual space representation, we can thus, e.g., enumerate which properties are satisfied by a given object, determine whether two concepts are disjoint or not, or rank objects according to a given (salient) ordinal feature.

Conceptual spaces were proposed as a framework for studying cognitive and linguistic phenomena, such as concept combination, metaphor and vagueness. As such, the problem of learning conceptual spaces from data has not received much attention. Within a broader setting, however, several authors have studied approaches for learning vector space representations that share important characteristics with conceptual spaces. The main focus in this context has been on learning vector space models with interpretable dimensions. For example, it has been proposed that non-negative matrix factorization leads to representations with dimensions that are easier to interpret than those obtained with other matrix factorization methods \cite{lee1999learning}, especially when combined with sparseness constraints \cite{hoyer2004non}. More recently, a large number of neural network models have been proposed with the aim of learning vectors with interpretable dimensions, under the umbrella term of \emph{disentangled representation learning} \cite{chen2016infogan,higgins2017beta}. Another possibility, advocated in \cite{derracAIJ}, is to learn (non-orthogonal) directions that model interpretable salient features within a vector space whose dimensions themselves may not be interpretable. 
Beyond interpretable dimensions, some authors have also looked at modelling properties and concepts as regions in a vector space. For example, \cite{Erk:2009:RWR:1596374.1596387} proposed to learn region representations of word meaning. More recent approaches along these lines include \cite{DBLP:journals/corr/VilnisM14}, where words are modelled as Gaussian distributions, and \cite{DBLP:conf/conll/JameelS17}, where word regions were learned using an ordinal regression model with a quadratic kernel. Some authors have also looked at inducing region based representations of concepts from the vector representations of known instances of these concepts \cite{DBLP:conf/ijcai/BouraouiS18,DBLP:journals/corr/abs-1808-01662}. Finally, within a broader setting, some approaches have been developed that link vectors to natural language descriptions, for instance linking word vectors to dictionary definitions \cite{hill2016learning} or images to captions \cite{karpathy2015deep}. 
 
The aforementioned approaches have found various applications. Within the specific context of explainable machine learning, at least two different strategies may be pursued. One possibility is to train a model in the usual way (e.g., a neural network classifier), and then approximate this model based on the semantic description of the vector representations involved. For instance, \cite{ager2016inducing} suggests a method for learning a rule based classifier that describes a feedforward neural network. The second possibility is to extract an interpretable (qualitative) representation from the vector space, e.g., by treating interpretable dimensions as ordinal features, and then train a model on that interpretable representation \cite{derracAIJ}. 
 
\subsection{Combining Deep Learning with High Level Inference}
While neural networks are traditionally learned in a purely data-driven way, several authors have explored approaches that are capable of taking advantage of symbolic knowledge. One common approach consists in relaxing symbolic rules, expressing available background knowledge, using fuzzy logic connectives. This results in a continuous representation of the background knowledge, which can then simply be added to the loss function of the neural network model \cite{DBLP:conf/emnlp/DemeesterRR16}. Rather than directly regularizing the loss function in this way, \cite{DBLP:conf/acl/HuMLHX16} proposes an iterative method to ensure that the proportion of ground instances of the given rules that are predicted to be true by the neural network is in accordance with the confidence we have in these rules. To this end, after each iteration, they solve an optimisation problem to find the set of predictions that is closest to the predictions of the current neural network while being in accordance with the rules. The neural network is subsequently trained to mimic these regularized predictions. Yes another approach is proposed in \cite{DBLP:conf/icml/XuZFLB18}, which proposes a loss function that encourages the output of a neural network to satisfy a predefined set of symbolic constraints, taking advantage of efficient weighted model counting techniques.

While the aforementioned approaches are aimed at using rules to improve neural network models, some authors have also proposed ways in which neural networks can be incorporated into symbolic formalisms. One notable example along these lines is the DeepProbLog framework from \cite{manhaeve2018deepproblog}, where neural networks are essentially used to define probabilistic facts of a probabilistic logic program. Within a broader setting, vector space embeddings have also been used for predicting plausible missing rules in ontological rule bases \cite{DBLP:conf/aaai/BouraouiS19,DBLP:conf/semweb/LiBS19}.


\subsection{Knowledge Graph Completion}
Knowledge graphs are a popular formalism for expressing factual relational knowledge using triples of the form (entity, relation, entity). In application fields such as natural language processing, they are among the most widely used knowledge representation frameworks. Such knowledge graphs are almost inevitably incomplete, however, given the sheer amount of knowledge about the world that we would like to have access to and given the fact that much of this knowledge needs to be constantly updated. This has given rise to a wide range of methods for automatic knowledge graph completion. On the one hand, several authors have proposed approaches for automatically extracting missing knowledge graph triples from text \cite{riedel2010modeling}. On the other hand, a large number of approaches have been studied that aim to predict plausible triples based on statistical regularities in the given knowledge graph. Most of these approaches rely on vector space embeddings of the knowledge graph \cite{bordes2013translating,trouillon2016complex}. The main underlying idea is to learn a vector $\mathbf{e}$ of typically a few hundred dimensions for each entity $e$, and a scoring function $s_R$ for each relation $R$, such that the triple $(e,R,f)$ holds if and only if $s_R(\mathbf{e},\mathbf{f})\in \mathbb{R}$ is sufficiently high. Provided that the number of dimensions is sufficiently high, any knowledge graph can in principle be modelled exactly in this way \cite{DBLP:conf/nips/Kazemi018}. To a more limited extent, such vector representations can even capture ontological rules \cite{DBLP:conf/kr/Gutierrez-Basulto18}. In practice, however, our aim is usually not to learn an exact representation of the knowledge graph, but to learn a vector representation which is predictive of triples that are plausible, despite not being among those in the given knowledge graph. Some authors have also proposed methods for incorporating textual information into knowledge graph embedding approaches. Such methods aim to learn vector space representations of the knowledge graph that depend on both the given knowledge graph triples and textual descriptions of the entities \cite{DBLP:conf/emnlp/ZhongZWWC15,DBLP:conf/ecai/JameelS16,xie2016representation,SSP}, or their relationships \cite{DBLP:conf/naacl/RiedelYMM13,toutanova2015representing}.


\subsection{Declarative Frameworks for Data Mining and Clustering}

Machine Learning and Data Mining can also be studied from the viewpoint of problem solving. From this point of view, two families of problems can be distinguished: enumeration and optimisation, the latter being either discrete or continuous. 

Pattern mining is the best known example of enumeration problems, with the search for patterns satisfying some properties, as for instance to be frequent, closed, emergent \ldots Besides, supervised classification is seen as the search for a model minimizing a given loss function, coupled to a regularization term for avoiding over-fitting, whereas unsupervised learning is modeled as the search of a set of clusters (a partition in many cases) optimizing a quality criterion (the sum of squared errors for instance for k-means). For sake of complexity optimisation problems usually rely on heuristic, with the risk of finding only a local optimum. All these approaches suffer from drawbacks. For instance in pattern mining the expert is often drowned under all the patterns satisfying the given criteria. In optimisation problems, a local optimum can be far from the expert expectations.

To prevent this, the notion of Declarative Data Mining has emerged, allowing the experts to express knowledge in terms of constraints on the desired models. It can be seen as a generalization of semi-supervised classification, where some points are already labelled with classes. Classical algorithms must then be adapted to take into account constraints and that has led to numerous extensions. Nevertheless, most extensions are dedicated to only one type of constraints, since the loss function has to be adapted to integrate their violation and the optimization method (usually a gradient descent) has to be adapted to the new optimization criterion. It has been shown in \cite{RaedtGN08} that declarative frameworks, namely Constraint Programming in that paper, allow to model and handle different kinds of constraints in a generic framework, with no needs to rewrite the solving algorithm. This has been applied to pattern mining and then extended to k-pattern set mining with different applications, such as conceptual clustering or tiling \cite{RaedtGN10,KhiariBC10}. 

This pioneering work has opened the way to a new branch of research, mainly in Pattern mining and in Constrained Clustering. In this last domain, the constraints were mainly pairwise, e.g., a Mustlink (resp. Cannotlink) constraint expresses that two points must (resp. must not) be in the same cluster. Other constraints have been considered such as cardinality constraints on the size of the clusters, minimum split constraints between clusters. Different declarative frameworks have been used, as for instance SAT \cite{DavidsonRS10,JabbourSS17}, CP \cite{DaoDV17}, Integer Linear Programming \cite{MuellerK10,BabakiGN14,KuoRDVD17,OualiZLLCBL17}. An important point is that such frameworks allow to easily embed symbolic and numerical information, for instance by considering a continuous optimisation criterion linked with symbolic constraints, or by considering two optimisation criteria and building a Pareto front \cite{KuoRDVD17}. 

Thus, declarative frameworks not only allow to easily integrate constraints in Machine Learning problems, but they enable the integration of more complex domain knowledge that goes beyond classical Machine Learning constraints, thus integrating truly meaningful conntraints \cite{DaoVDD16}. Moreover, new use case for clustering can be considered as for instance given a clustering provided by an algorithm, find a new clustering satisfying new expert knowledge, modifying a minima the previous clustering. The price to pay is computational complexity, and the inability to address large datasets. A new research direction aims at studying how constraints could be integrated to deep learner \cite{corr/abs-1901-10061}. 
Besides, the Constraint Programming community has also benefited from this new research direction by the development of global constraints tailored to optimize the modeling of Data Mining tasks, as for instance \cite{KemmarLLBC17}.


%

\input{MLvsAR}

\subsection{Preferences and Recommendation}

 
In both KRR and ML, models have evolved from 
binary ones (classical propositional or first-order logic, binary classification) to richer ones that take into account the need to propose less drastic outputs. One approach has been to add the possibility to \emph{order} possible outputs / decisions. In multi-class classification tasks for instance, one approach \cite{DuanKeerthiChuShevadePoo:mcs03} is to estimate, given an instance, the \textit{posterior} probability of belonging to each possible class, and predict the class with highest probability. The possibility of learning to ``order things'' has numerous applications, e.g., in information retrieval, recommender systems. In KRR, the need to be able to order interpretations (rather than just classify them as possible / impossible, given the knowledge at hand) has proved to be an essential modelling paradigm, see, e.g., the success of valued CSPs \cite{SchiexFargierVerfaillie:ijcai95}, Bayesian networks \cite{Pearl:88}, possibilistic / fuzzy logics among others.

At the intersection of ML and KRR, the field of ``preference learning'' has emerged. Furnkranz {et al.} \cite{FurnkranzH10,FurnkranzHRSS14} describe various tasks that can be seen as preference learning, some where the output is a function that orders possible labels for each unseen instance, and some where the output is a function that orders any unseen set of new instances.

\medskip

The importance of the notion of preference seems to have emerged first in economics and decision theory, and research in these fields focused essentially on \textit{utilitarian} models of preferences, where utility function associates a real number with each one of the objects to be ordered. Also in this field research developed on \emph{preference elicitation}, where some interaction is devised to help a decision maker form / lay bare her preferences, usually over a relatively small set of alternatives, possibly considering multiple objectives.

In contrast, preferences in AI often bear on combinatorial objects, like models of some logical theory to indicate for instance preferences over several goals of an agent; or, more recently, like combinations of interdependent decisions or configurable items of some catalog.
Thus, in KRR as well as in ML, the objects to be ordered are generally characterised by a finite number of features, with a domain / set of possible values for each feature. When talking about preferences, the domains tend to be finite ; continuous domains can be discretised.

Because of the combinatorial nature of the space of objects, research in AI emphasized the need for \emph{compact} models of preferences. Some probabilistic models, like Bayesian networks or Markov random fields, fall in this category, as well as e.g., additive utilities \cite{Fishburn:IntEcRev67} and their generalisations. This focus on combinatorial objects also brought to light one difficulty with the utilitarian model: although it is often easy to compute the utilities or probabilities associated with two objects and compare them on such a basis, it appears to be often NP-hard to find optimal objects from a combinatorial set with numerical representations of preferences. 
Thus one other contribution of research in KRR is to provide preference representation languages where optimisation is computationally easy, like CP-nets \cite{Boutilieretal:jair04}.

These complex models of preferences have recently been studied from an ML perspective, both in an elicitation / active learning setting, and in a batch / passive learning setting. One particularity of these compact preference models is that they combine two elements: a structural element, indicating probabilistic or preferential interdependencies between the various features characterizing the objects of interest; and ``local'' preferences over small combinations of features. It is the structure learning phase which is often demanding, since finding the structure that best fits some data is often a hard combinatorial search problem. In contrast, finding the local preferences once the structure has been chosen is often easy.

The passive learning setting is particularly promising because of the vast dataset available in potential applications of preference learning in some decision aid systems like recommender systems or search engines. The possibility to learn Bayesian networks from data has been a key element for their early success in many applications. Note that in some applications, in particular in the study of biological systems, learning the structure, that is, the interdependencies between features, is interesting; in such applications, ``black-box'' models like deep neural networks seem less appropriate. This is also the case in decision-support systems where there is a need to \emph{explain} the reasons justifying the computed ordering of possible decisions \cite{BelahceneLabreucheMaudetMousseauOuerdane_th-dec17}.

At the frontier between learning and reasoning lies what could be named lazy preference learning: given a set of preference statements which do not specify a complete preference relation, one can infer new pairwise comparisons between objects, by assuming some properties the full, unknown preference relation. As a baseline, many settings in the models studied in KRR assume transitivity of preferences, but this alone does not usually induce many new comparisons. A common additional assumption, made by \cite{BelahceneLabreucheMaudetMousseauOuerdane_th-dec17}, is that the preference relation can be represented with an additive utility function, and that the ordering over the domain of each feature is known. In \cite{Wilson:ecai06,Wilson:ijcai09,Wilson:ecai14}, richer classes of input preference statements are considered, and the assumption is made that the preference relation has some kind of (unknown) lexicographic structure.

Lastly, as mentioned above, analogical proportions can be used for predicting preferences \cite{FH18,BounhasPP18,BounhasPPS19}. The general idea is that a preference between two items can be predicted if some analogical proportions hold that link their descriptions with the descriptions of other items for which preference relations are known.

\section{Conclusion}
The KRR and ML areas of AI have been developed independently to a large extent for several decades. It results that most of the researchers in one area are ignorant of what has been going on in the other area. The intended purpose of this joint work is to provide an inventory of the meeting points between KRR and ML lines of research. We have first reviewed some concerns that are shared by the two areas, maybe in different ways. Then we have surveyed various paradigms that are at the border between KRR and ML. Lastly, we have given an overview of different hybridizations of KRR and ML tools. 

Let us emphasize that this is a work in progress. Subsections may be at this step unequally developed, and remain sketchy. The current list of references is certainly incomplete and unbalanced. The works covered may be old or recent, well-known as well as overlooked. 
However, at this step, we have absolutely no claim of completeness of any kind, not even of being fully up to date. Examples of topics not covered at all are numerous. They include reinforcement learning, argumentation and ML \cite{AmgoudS08}, ontology representation and learning \cite{onto15}, in relation with the modeling of concepts, rough sets and rule extraction \cite{Pawlak91,Grzymala-BusseZ00}, or formal studies of data such as version space learning \cite{Mitchell} or logical analysis of data
\cite{BorosCHIKM11}; see also \cite{3approaches,Mirkin}.

Moreover, each topic covered in this paper is only outlined and would deserve to be discussed in further details. This paper is only a first step towards a more encompassing and extensive piece of work. 

The aim of this paper is to help facilitating the understanding between researchers in the two areas, with a perspective of cross-fertilisation and mutual benefits. Still we should be aware that the mathematics of ML
and the mathematics of KRR are quite different if we consider the main trends in each area. In ML the basic paradigm is a matter of approximating functions (which then calls for optimization). The mathematics of ML are close to those of signal processing and automatic control 
(as pointed out in \cite{lecun19}), while KRR is dominated by logic and discrete mathematics, leading to an -- at least apparent -- opposition between geometry and logic \cite{Mallat18}\footnote{See also the conference https://www.youtube.com/watch?v=BpX890StRvs}. But functions also underlie KRR, once one notices that a set of (fuzzy or weighted) rules is like an aggregation function \cite{DubPra97}, whose computation may take the form of a generalized matrix calculus (`generalized' in the sense that the operations are not necessarily restricted to sum and product). Let us also note that the convolution of functions (a key tool in signal processing) is no longer restricted to a linear, sum/product-based setting \cite{iwann/MolekP19}. 

Besides, it seems difficult to envisage ML without KRR. For instance, it has been recently observed that
\begin{itemize}
    \item the unsupervised learning of disentangled representations is fundamentally impossible without inductive biases on both the models and the data \cite{Locatello};
    \item local explanation methods for deep neural networks lack sensitivity to parameter values 
\cite{AdebayoGGK18};
\item when trained on one task, then trained on a second task, many machine learning models ``forget'' how to perform the first task 
\cite{GoodfellowMDCB14}.
\end{itemize}  
 Such states of fact might call for some cooperation in the long range between ML and KRR. 

\section{Acknowledgements} 
The authors thank the CNRS GDR ``Formal and Algorithmic Aspects of Artificial Intelligence'' for its support. 

\bibliography{ref_sum,MLvsAR} 

\end{document}

%% file: MLvsAR.tex
\subsection{Machine Learning vs.~Automated Reasoners} \label{sect:sat}

This subsection provides a brief overview of the applications of ML in practical Automated Reasoners, but also overviews the
recent uses of Automating Reasoning in ML\footnote{%
We adopt a common understanding of \emph{Automated Reasoning} as
``\emph{The study of automated reasoning helps produce computer
  programs that allow computers to reason completely, or nearly
  completely, automatically}'' (from
\url{https://en.wikipedia.org/wiki/Automated_reasoning}).}. With a few
exceptions, the subsection emphasizes recent work, published over the
last decade.
Tightly related work, e.g.\ inductive logic
programming (see Section \ref{ILP}) or statistical relational
learning~\cite{deraedt-bk16}, is beyond the scope of this subsection.
The subsection is organized in three main parts. First, \autoref{sec:L4R}
overviews the applications of ML in developing and organizing
automated reasoners. Second, \autoref{sec:R4L} covers the recent users
of automated reasoning in learning ML models, improving the robustness
of ML models, but also in explaining ML models.
Finally, \autoref{sec:LvsR} covers a number of recent topics at the
intersection of automated reasoning and ML.
%


\subsubsection{Learning for Reasoning} \label{sec:L4R}

Until recently, the most common connection between ML
and automated reasoning would be to apply the former when devising
solutions for the latter.
As a result, a wealth of attempts have been made towards applying ML
in the design of automated reasoners, either for improving existing
algorithms or for devising new algorithms, built on top of ML models.
Uses of ML can be organized as follows. First, uses of ML for
improving specific components of automated reasoners, or for automatic
configuration or tuning of automated reasoners. Second, approaches
that exploit ML for solving computationally hard decision, search and
counting problems, and so offering alternatives to dedicated automated
reasoners.

\paragraph{Improving Reasoners.}
Earlier efforts on exploiting ML in automated reasoners was to improve
specific components of reasoners by seeking guidance from some ML
model.
A wealth of examples exist, including the improvement of restarts in
Boolean Satisfiability (SAT) solvers~\cite{walsh-sat09}, improvement
of branching
heuristics~\cite{blaschko-nips12,popescu-fi14,ganesh-sat16,ganesh-aaai16,ganesh-sat18},
selection of abstractions for Quantified Boolean Formulas (QBF)
solving~\cite{janota-aaai18,seshia-corr18}, but also for improving
different components of theorem provers for first-order and higher
order
logics~\cite{urban-tableaux11,urban-jar14,urban-paar14,urban-jsc15,szegedy-nips16,szegedy-lpar17,urban-cade19}.

ML has found other uses for improving automated reasoners. A
well-known example is the organization of portfolio
solvers~\cite{hoos-jair08,hoos-lion11,hoos-aij14,hoos-aij16,mplms-ecai08}.
Another example is the automatic configuration of solvers, when the
number of options available is large~\cite{hoos-jair09,ganesh-tr09}.
One additional example is the automatic building of automated
reasoners using ML~\cite{hoos-aij16}.


\paragraph{Tackling Computationally Hard Problems.}

Another line of work has been to develop solutions for solving
computationally hard decision and search problems.
Recent work showed promise in the use of NNs for solving satisfiable
instances of SAT represented in clausal
form~\cite{selsam-corr18,selsam-iclr19,selsam-sat19,kolter-icml19},
for solving instances of SAT represented as
circuits~\cite{weimer-iclr19,weimer-corr19}, but also NP-complete
problems in general~\cite{vardi-aaai19}.
The most often used approach has been to exploit variants of Graph
Neural Networks~(GCNs)~\cite{scarselli-ieee-tnn09}, including
Message Passing Neural Networks~(MPNNs)~\cite{vinyals-icml17}.
There has also been recent work on solving CSPs~\cite{koenig-cp18}
using convolutional NNs. Furthermore, there have been proposals for
learning to solve SMT~\cite{vechev-nips18a}, combinatorial
optimization
problems~\cite{dilkina-nips17,bengio-iclrw17,koltun-nips18,bengio-corr18},
planning problems~\cite{winther-gcai18},
but also well-known specific cases of NP-complete decision problems,
e.g.\ Sudoku~\cite{winther-nips18} and TSP~\cite{vinyals-nips15}.

Efforts for tackling computationally harder problems have also been
reported, including QBF~\cite{yang-corr19b}, ontological
reasoning~\cite{lukasiewicz-corr17}, probabilistic logic
programming~\cite{deraedt-nips18}, inference in probabilistic
graphical models~\cite{zemel-iclrw18b} and theorem
proving for first
order~\cite{urban-nips18,huang-iclr19,szegedy-icml19,szegedy-corr19,huang-corr19}
and higher order
logics~\cite{whalen-corr16,deng-nips17,szegedy-iclr17,deng-icml19}.

\subsubsection{Reasoning for Learning} \label{sec:R4L}

This subsection overviews uses of automated reasoning approaches for
verifying, explaining and learning ML models.

\paragraph{Robust Machine Learning.}

Concerns about the behavior of neural networks can be traced at least
to the mid 90s and early
00s~\cite{napolitano-waias96,zakrzewski-ijcnn01,schumann-woss02}.
Additional early work on ensuring safety of neural networks also
involved SAT solvers~\cite{pulina-cav10}.
More recently, efforts on the verification of neural networks have
focused on the avoidance of so-called adversarial examples.

Adversarial examples~\cite{szegedy-iclr14}, already briefly mentioned in Subsection \ref{causex}, illustrate the brittleness
of ML models. In recent years, a number of unsettling examples
served to raise concerns on the fragility neural networks can be in
practice~\cite{aung-corr17,song-cvpr18,chakraborty-corr18,kohane-science19,heaven-nature19}. 
Among other alternative approaches, automated reasoners have been
applied to ensuring the robustness of ML models, emphasizing neural
networks.
A well-known line of work focuses on the avoidance of adversarial
examples for neural networks using ReLU units~\cite{hinton-icml10} and
proposes Reluplex, an SMT-specific dedicated reasoning engine for
implementing reasoning with ReLU
units~\cite{barrett-cav17,barrett-atva18,barrett-cav19}.
Another recent line of work addresses binarized neural
networks~\cite{bengio-nips16} and develops a propositional encoding
for assessing the robustness of
BNNs~\cite{narodytska-aaai18,narodytska-ijcai18,narodytska-corr18}.
Additional lines of work have been reported in the last two
years~\cite{kwiatkowska-cav17,kwiatkowska-ijcai18,kwiatkowska-ijcai19,kwiatkowska-concur19,vechev-sp18,vechev-icml18,vechev-nips18b,vechev-iclr19,vechev-popl19,vechev-corr19,kohli-uai18,kohli-nips18,kohli-corr18,kohli-iclr19c,kohli-uai19,kohli-cvpr19,kumar-corr17,kumar-iclr19,kumar-corr19,seshia-atva18,seshia-corr19,meel-ccs19}.

\paragraph{Interpretable ML Models.}
Interpretable ML models are those from which rule-like explanations
can be easily produced. For example, decision trees, decision sets (or
rule sets) and rule lists are in general deemed interpretable, since
one can explain predictions using rules.
On area of research is the learning (or synthesis) of interpretable ML
models using automated reasoning approaches.
There has been continued efforts at learning decision
trees~\cite{nijssen-kdd07,nijssen-icml08,bessiere-cp09,nijssen-dmkd10,verwer-cpaior17,nipms-ijcai18,verwer-aaai19,schaus-cj19,rudin-nips19},
decision sets~\cite{leskovec-kdd16,ipnms-ijcar18,meel-cp18,meel-aies19}
and rule lists~\cite{rudin-kdd17,rudin-jmlr17}. Examples of reasoners
used include SAT, CP, and ILP solvers, but dedicated complete methods
based on branch and bound search have also been considered.
Despite a recent explosion of works on black-box ML models, there exist
arguments for the use of interpretable models~\cite{rudin-nmi19}.

\paragraph{Explanations with Abductive Reasoning.}
In many settings, interpretable models are not often the option of
choice, being replaced by so-called black-box models, which include 
any ML model from which rules explaining predictions are not readily
available.~\footnote{%
  The definition of \emph{explanation} is the subject of ongoing
  debate~\cite{miller-aij19}. We use the intuitive notion of
  explanation as a {IF-THEN}
  rule~\cite{guestrin-kdd16,lundberg-nips17,guestrin-aaai18}, where
  some given prediction is made if a number of feature 
  values holds true. The importance of reasoning about explanations is
  illustrated by a growing number of recent
  surveys~\cite{klein-ieee-is17a,klein-ieee-is17b,cotton-ijcai07-xai,muller-dsp18,klein-ieee-is18a,klein-ieee-is18b,berrada-ieee-access18,mencar-ipmu18,hlupic-mipro18,klein-corr18,pedreschi-acmcs19,xai-bk19,muller-xai19-ch01,miller-aij19,miller-acm-xrds19,anjomshoae-aamas19,russell-fat19a,zhu-nlpcc19,klein-corr19}.} 
Concrete examples include (deep) neural networks (including
binarized versions), and boosted trees and random forests, among many
other alternatives.

Most existing works on computing explanations resort to so-called
\emph{local} explanations. These models are agnostic and heuristic in
nature~\cite{guestrin-kdd16,lundberg-nips17,guestrin-aaai18}. Recent
works~\cite{nsmims-sat19,inms-corr19} revealed that local explanations
do not hold globally, i.e., it is often the case that there are points
in feature space, for which the local explanation holds, but for which
the model's prediction differs.
Since 2018, a number of attempts have been reported, which propose
rigorous approaches for computing explanations. Concretely, these
recent attempts compute so-called \emph{abductive} explanations, where
each explanation corresponds to a prime implicant of the discrete
function $\mathcal{F}$ representing the constraint that the ML model
predicts the target prediction.
A first attempt based on compiling such a  function into a tractable
representation is reported elsewhere~\cite{darwiche-ijcai18}. For such a
representation, (shortest) prime implicants can then be extracted in
polynomial time. The downside of this approach is that compilation may
yield exponential size function representations.
Another attempt~\cite{inms-aaai19} is based on computing explanations
of demand, by encoding the instance, the ML model and the prediction
into some logic representation. In this case, reasoners such as SMT,
ILP or SAT solvers are then used for extracting (shortest) prime
implicants.
%

\paragraph{Explanations vs.\ Adversarial Examples.}
In recent years, different works realized the existence of some
connection between adversarial examples (AE's) and explanations
(XP's)~\cite{hu-kdd18,zhang-nips18,doshi-velez-aaai18,tomsett-fusion18,lin-corr18a,roy-corr18a,wu-corr18a}.
Nevertheless, a theoretical connection between AE's and XP's has been
elusive. Recent work~\cite{inms-aaai19} showed that adversarial
examples can be computed from the set of explanations for some
prediction.
Furthermore, this work introduced the concept of counterexample (CEx)
to some prediction, and identified a minimal hitting set relationship
between XP's and CEx's, i.e., XP's are minimal hitting sets of CEx's
and vice-versa.
%

\subsubsection{More on Learning~vs.~Reasoning} \label{sec:LvsR}

The previous two subsections summarize recent efforts on using machine 
learning for automated reasoning, but also on using automated
reasoning for learning, verifying and explaining ML 
models.
This subsection identifies additional lines of research at the
intersection of ML and automated reasoning.

%

\paragraph{Integrating Logic Reasoning in Learning.}
A large body of work has been concerned with the integration of logic
reasoning with ML. One well-known example is
neural-symbolic
learning~\cite{garcez-bk02,garcez-bk09,garcez-corr17,kohli-iclr17,kohli-icse18,kohli-iclr19b,garcez-corr19}. See also Subsection \ref{neurosy}. Examples
of applications include program
synthesis~\cite{zemel-iclrw18a,zemel-nips18,kohli-iclr17,kohli-iclr18b} 
and neural theorem proving~\cite{riedel-corr18}.
Other approaches do exist, including deep reasoning
networks~\cite{gomes-corr19}, neural logic machines~\cite{zhou-iclr19},
and abductive learning~\cite{zhou-corr18,zhou-sc19}.
An alternative is to embed symbolic knowledge in neural
networks~\cite{meel-corr19}.

\paragraph{Learning for Inference.}
One area of work is the use of ML models for learning logic
representations, most often
rules~\cite{riedel-akbc16,riedel-nips17,cohen-nips17,grefenstette-ijcai18,grefenstette-jair18},
which can serve for inference or for explaining predictions.

\paragraph{Understanding Logic Reasoning.}
A natural question is whether ML systems understand
logical formulas in order to decide entailment or unsatisfiability.
There has been recent work on understanding
entailment~\cite{kohli-iclr18a,kohli-iclr19a}, suggesting that this is
not always the case, e.g., for convolutional NNs. In a similar
fashion, recent work~\cite{yang-corr19a} suggests that GNNs mail fail
at deciding unsatisfiability.

\paragraph{Synthesis of ML Models.}
Recent work proposed the use of automated reasoners for the synthesis
(i.e., learning) of ML models. Concrete examples
include~\cite{vechev-icml19,jagannathan-pldi19,jagannathan-corr19}.
These approaches differ substantially from approaches for the synthesis
of interpretable models, including decision trees and sets and
decision lists.\\


As witnessed by the large bibliography surveyed in this subsection, the
quantity, the breadth and the depth of existing work at the
intersection between ML and automated reasoning in recent years,
provides ample evidence that this body of work is expected to continue
to expand at a fast pace in the near future.